\documentclass[10pt]{wlscirep}
\usepackage[utf8]{inputenc}
\usepackage[T1]{fontenc}
\usepackage{diagbox}   
\usepackage{booktabs}  
\usepackage{multirow}  
\usepackage{makecell}
\usepackage[numbers,super,sort&compress]{natbib}
\usepackage{url}
\usepackage{graphicx}
\usepackage{hyperref}

\title{EPBC-YOLOv8: An efficient and accurate improved YOLOv8 underwater detector based on an attention mechanism}

\author[1]{Xing Jiang}
\author[1]{Xiting Zhuang}
\author[1]{Jisheng Chen}
\author[1,*]{Jian Zhang}
\author[1]{Yiwen Zhang}
\affil[1]{School of Tropical Agricultureand Forestry(School of Agricultural and Rural, School of Rural Revitalization), Hainan University, Haikou 570228, China}

\affil[*]{whealther@hainanu.edu.cn}

\begin{abstract}
In this study, we enhance underwater target detection by integrating channel and spatial attention into YOLOv8's backbone, applying Pointwise Convolution in FasterNeXt for the FasterPW model, and leveraging Weighted Concat in a BiFPN-inspired WFPN structure for improved cross-scale connections and robustness. Utilizing CARAFE for refined feature reassembly, our framework addresses underwater image degradation, achieving mAP@0.5 scores of 76.7\% and 79.0\% on URPC2019 and URPC2020 datasets, respectively. These scores are 2.3\% and 0.7\% higher than the original YOLOv8, showcasing enhanced precision in detecting marine organisms.

\end{abstract}

\begin{document}

\flushbottom
\maketitle
\thispagestyle{empty}

\section*{Introduction}

The advancement of machine vision technology has enabled underwater robots and aerial vehicles to utilize non-invasive, high-resolution visual perception for autonomous exploration and exploitation of marine resources. Detecting and identifying marine life is crucial for utilizing these resources. However, challenges such as low underwater image quality, small and clustered targets, and imbalanced quantities pose difficulties for this task. Existing methods, primarily based on generic object detection techniques, have been applied in aquaculture\cite{Multi-stream fish detection in unconstrained underwater videos by the fusion of two convolutional neural network detectors}, distant-water fisheries\cite{Potential Efficiency of Earth Observation for Optimum Fishing Zone Detection of the Pelagic Sardinella Aurita Species along the Mediterranean Coast of Egypt}, and marine species monitoring\cite{Projecting future changes in distributions of small-scale pelagic fisheries of the southern Colombian Pacific Ocean}. Nevertheless, underwater object detection (UOD) faces issues like low image quality, small and clustered targets, and computational limitations, leading to suboptimal results. There is a need for suitable deep learning models to improve the accuracy and efficiency of UOD, but currently, few researchers focus on addressing these problems.

However, the harsh underwater environment often results in issues such as noise \cite{A Marine Organism Detection Framework Based on the Joint Optimization of Image Enhancement and Object Detection}, uneven lighting conditions\cite{liEnhancing2022}, blurriness \cite{Enhanced Frequency Fusion Network with Dynamic Hash Attention for Image Denoising}, and low contrast\cite{Attention Guided Low-Light Image Enhancement with a Large Scale Low-Light Simulation Dataset 2020}, which degrade the performance of traditional object detection models like YOLOv8. To address this, attention mechanisms have been increasingly incorporated into detection frameworks, enhancing the model's focus on relevant features. The interaction across dimensions in attention mechanisms helps the model understand relationships between different features, improving detection performance. However, traditional attention mechanisms may face computational challenges with large-size images.  Ouyang et al.\cite{Efficient Multi-Scale Attention Module with Cross-Spatial Learning} proposed a new approach, the Multi-scale Attention Module (EMA), introduces cross-dimensional interaction to better handle features of different scales, offering potential improvements in channel or spatial attention\cite{Recurrent Models of Visual Attention} prediction tasks.

Underwater target detection faces challenges due to small and clustered organisms against complex backgrounds. Traditional single-stage object detection algorithms struggled with representing multi-scale objects, leading to the evolution of feature pyramid network (FPN)\cite{Feature Pyramid Networks for Object Detection. In Proceedings of the 2017 IEEE Conference on Computer Vision and Pattern Recognition (CVPR)}  algorithms that utilize multi-stage feature maps. However, FPN overlooks the varying significance of features at different levels during fusion. To address this, studies like NAS-FPN \cite{NAS-FPN: Learning Scalable Feature Pyramid Architecture for Object Detection}and BiFPN\cite{EfficientDet: Scalable and Efficient Object Detection} introduced irregular feature fusion modules and weighted feature fusion, respectively. Inspired by BiFPN's cross-scale connections and weighted feature fusion, we employ Weighted\_Concat for fusion and introduce WFPN to enhance cross-scale connections, enabling better integration of positional and detailed information.

The complex underwater environment necessitates larger receptive fields and semantic association for enhanced detection performance. Traditional interpolation-based upsampling methods fall short as they increase resolution without adding feature information. This study adopts the CARAFE \cite{CARAFE: Content-Aware ReAssembly of FEatures} upsampling technique, which expands the receptive field and aggregates contextual information effectively, while being lightweight with fewer parameters and computational demands. This approach holds promise for improving underwater feature map understanding and detection performance.

This paper presents an underwater object detector based on YOLOv8\cite{YOLOv8 by Ultralytics}, with several enhancements: 

•  Integration of the EMA multi-scale attention module with YOLOv8's C2f backbone, improving responsiveness to targets of different scales and reducing feature redundancy. 

•  Enhancement of YOLOv8 with the FasterNext \cite{chenRun2023} module and replacement of Partial Convolution with Pointwise Convolution (PWConv) in the proposed FasterPW model, boosting the backbone's lightweight nature and feature extraction capability. 

•  Utilization of WFPN's cross-scale connections and weighted feature fusion for more effective integration of positional and detailed information. 

•  Replacement of the original upsampling module with CARAFE, a content-aware strategy, to preserve the model's ability to extract information from small targets without losing detail features due to interpolation-based upsampling. 

The proposed EPBC-YOLOv8 underwater object detector achieves a balance between computational efficiency and accuracy. On the URPC2019 and URPC2020 datasets, it attains mAP@0.5 scores of 76.7\% and 79.0\%, respectively, surpassing the original YOLOv8 by 2.3\% and 0.7\%.

The paper is structured as follows: Section "\hyperref[sec2]{Related Work}" reviews related work, Section "\hyperref[sec3]{EPBC-YOLOv8}" details the EPBC-YOLOv8 model, Section "\hyperref[sec4]{Experimental Details}" presents experimental analysis, and Section "\hyperref[sec5]{Conclusions and Future Work}" concludes and outlines future work.

\section*{Related Work}\label{sec2}

Object detection in computer vision is primarily categorized into one-stage and two-stage techniques. Two-stage methods, exemplified by R-CNN\cite{Rich Feature Hierarchies for Accurate Object Detection and Semantic Segmentation} and Faster R-CNN\cite{Faster R-CNN: Towards Real-Time Object Detection with Region Proposal Networks}, generate candidate bounding boxes and subsequently classify and refine them, achieving high accuracy at the expense of speed. Enhancements such as the integration of adversarial networks have been introduced to bolster robustness and expedite detection. In contrast, the main one-stage object detection algorithms include the YOLO family, which consists of YOLO \cite{redomnYOLO2016} that overcomes the shortcomings of two-stage detection networks. YOLOv2 \cite{YOLO9000} introduces batch normalization layers after each convolution and eliminates the use of dropout. YOLOv3 \cite{YOLOv3: An Incremental Improvement} marked a significant improvement, characterized by the introduction of the residual module Darknet-53 and FPN. There have been studies that added many techniques based on YOLOv3, such as YOLOv4 \cite{Yolov4: Optimal speed and accuracy of object detection}, YOLOv5 \cite{YOLOv5 by Ultralytics}, YOLOv6 \cite{YOLOv6: A Single-Stage Object Detection Framework for Industrial Applications 2022}, and YOLOv7 \cite{YOLOv7: Trainable Bag-of-Freebies Sets New State-of-the-Art for Real-Time Object Detectors}. The related code for YOLOv8 can be found on GitHub. SSD \cite{SSD: Single Shot MultiBox Detector} and RetinaNet \cite{Focal Loss for Dense Object Detection} are also part of this category. YOLO divides the entire image into a grid, with each cell predicting bounding boxes and classification confidence. Some studies have introduced modifications to YOLOv8 to achieve high-precision detection performance \cite{Underwater Object Detection Using TC-YOLO with Attention Mechanisms, Fish Detection under Occlusion Using Modified You Only Look Once v8 Integrating Real-Time Detection Transformer Features, Student Behavior Detection in the Classroom Based on Improved YOLOv8}

In computer vision, attention mechanisms such as channel attention, spatial attention, and combined channel-spatial attention dynamically adjust the weights of input features to focus on important areas, similar to the human visual system. Examples include CBAM\cite{CBAM: Convolutional Block Attention Module} and EMA for combined channel-spatial attention. The residual attention network\cite{Residual Attention Network for Image Classification}, an early implementation, generates a three-dimensional attention map but faces challenges with high computational cost and limited receptive fields. To improve efficiency, techniques like global average pooling and decoupling methods have been introduced. Additionally, the Feature Pyramid Network (FPN) constructs a hierarchical feature pyramid to represent objects of various sizes, enhancing multi-scale target detection by merging features from different network levels, thereby improving detection accuracy and capturing fine details. In underwater object detection, upsampling is crucial for reorganizing features to achieve higher detection performance. Methods like linear interpolation and deep learning-based upsampling, such as Meta-Upscale\cite{Meta-SR: A Magnification-Arbitrary Network for Super-Resolution} and CARAFE. While interpolation methods can increase image resolution, they may introduce noise and increase computational complexity. In contrast, deep learning-based methods like CARAFE offer a large receptive field and accurate detail restoration without significantly increasing computational complexity, making them effective for balancing detection performance improvement in the field of computer vision.

\section*{EPBC-YOLOv8}\label{sec3}

\begin{figure}[htbp]
    \centering
    \includegraphics[width=0.5\linewidth]{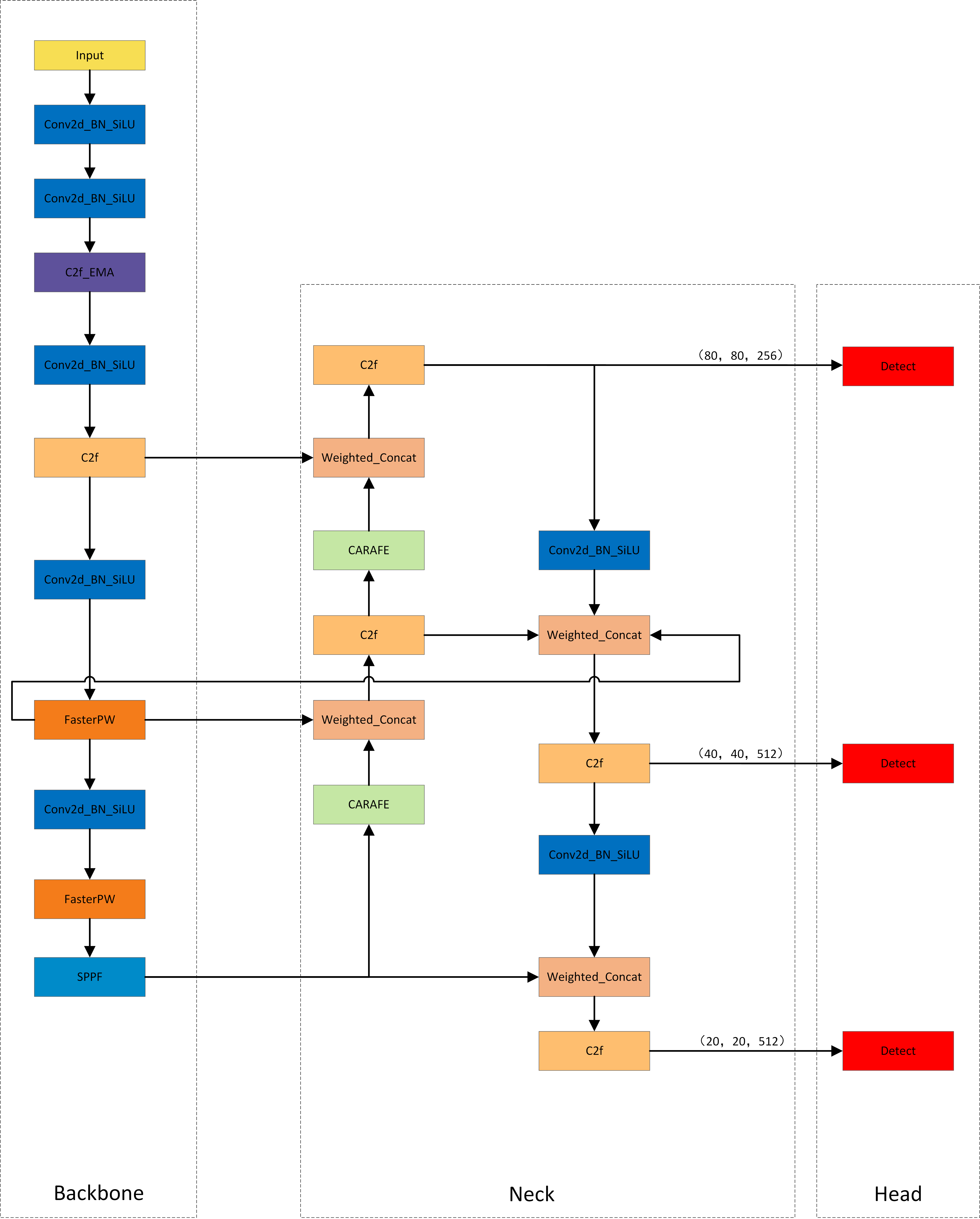}
    \caption{The structure of EPBC-YOLOv8.}
    \label{fig1:enter-label}
\end{figure}

Although the YOLOv8 model has achieved remarkable results in the field of object detection, it has some limitations. Firstly, the model's memory consumption and computational complexity are relatively high, limiting its deployment efficiency on resource-constrained edge devices. Secondly, the performance of YOLOv8 in detecting small objects needs improvement, especially in the detection of densely arranged small objects, where the model struggles to effectively learn feature information. Moreover, the robustness of the model in handling images with complex backgrounds needs to be enhanced. To address these issues, we designed EPBC-YOLOv8, which integrates C2f\_EMA, FasterPW, WFPN, and CARAFE into the YOLOv8 architecture, as shown in Figure.\ref{fig1:enter-label}.
\subsection*{C2f\_EMA}

The EMA mechanism is an innovative parallel processing framework designed specifically for computer vision tasks to enhance model performance and accelerate data processing speed. Its framework can be summarized into the following main parts:

•  Parallel Structure: EMA employs a parallel architecture to process input data, differing from the sequential layer structure of traditional Convolutional Neural Networks (CNNs). This parallel structure improves the efficiency of model training and enhances the model's accuracy when dealing with multi-scale features.

•  Feature Map Grouping: EMA groups input feature maps along the channel dimension, with each group processing a subset of features. This grouping strategy enhances the model's ability to handle different features and promotes model learning by assigning weights to different input features using attention mechanisms \cite{Cattle Body Detection Based on YOLOv5-EMA for Precision Livestock Farming}.

•   Multi-scale Spatial Information Capture: EMA captures multi-scale spatial information through parallel subnetworks with large receptive fields, enabling simultaneous processing of feature regions of different sizes. This approach allows for a more effective understanding and representation of various aspects of the input data.

•   Attention Weight Extraction: EMA is designed with three parallel paths to extract attention weights, including two 1×1 branches and one 3×3 branch. This configuration encodes information from different spatial directions and captures more complex multi-scale features.

•   Feature Interaction and Spatial Attention Map Generation: EMA processes interactions between different features through a cross-spatial information aggregation strategy, generating spatial attention maps. This enables spatial information of different scales to be effectively integrated within the same processing stage.

•   Final Output: The output of EMA includes two spatial attention maps, preserving precise spatial location information. The output feature maps within each group are further processed through a Sigmoid function to optimize the final feature representation.

\begin{figure}[htbp]
    \centering
    \includegraphics[width=0.6\linewidth]{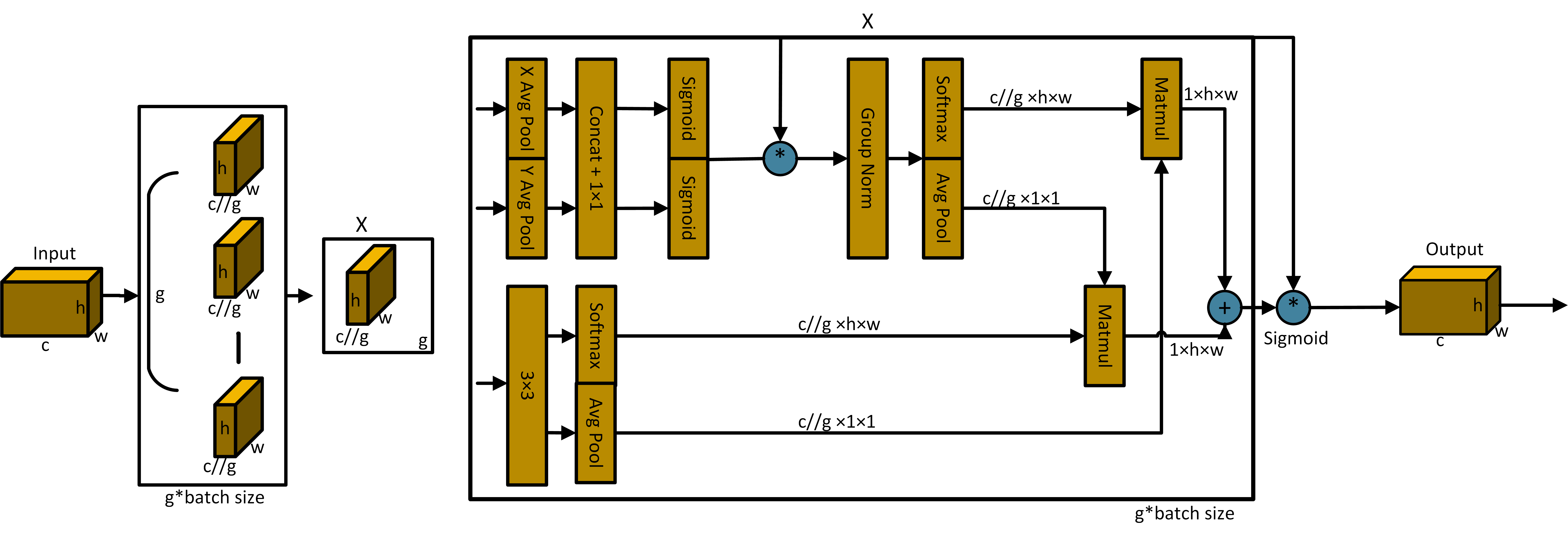}
    \caption{Schematic diagram of EMA. Here, 'g' represents grouping, 'X Avg Pool' represents 1D horizontal global pooling, and 'Y Avg Pool' represents 1D vertical global pooling.}
    \label{fig2:enter-label}
\end{figure}

In this study, we incorporated the EMA module into the neck part of YOLOv8, which is a key component of its improved version. As shown in Figure.\ref{fig2:enter-label}, EMA splits the input feature maps \(X \in \mathbb{R}^{C \times H \times W}\)into G groups of cross-channel sub-features, each group \(X_i \in \mathbb{R}^{C/G \times H \times W}\)learning different semantic information. To enhance the ability to capture multi-scale spatial information, we replaced the original 3×3 branch with a 5×5 branch, expanding the model's receptive field. EMA consists of three parallel paths, with two in the 1×1 branch and one in the 3×3 branch. Specifically, global spatial information is extracted from the output of the 1×1 branch using two-dimensional global average pooling. Meanwhile, the output from the 3×3 branch undergoes direct adjustment to align with the corresponding dimensional structure before the joint activation mechanism that incorporates channel features, as shown in Equation.(\ref{equ1}). With these improvements, the EMA module provides YOLOv8 with more efficient spatial information processing capabilities.
\begin{equation}
z_c=\frac{1}{H\times W}\sum_{j}\sum_{i} x_c\left(i,j\right)
\label{equ1}
\end{equation}
Here, \(z_c\) represents the output associated with the c-th channel. The primary purpose of this output is to encode global information, thereby capturing and modeling long-range dependencies.

\begin{figure}[htbp]
    \centering
    \includegraphics[width=0.5\linewidth]{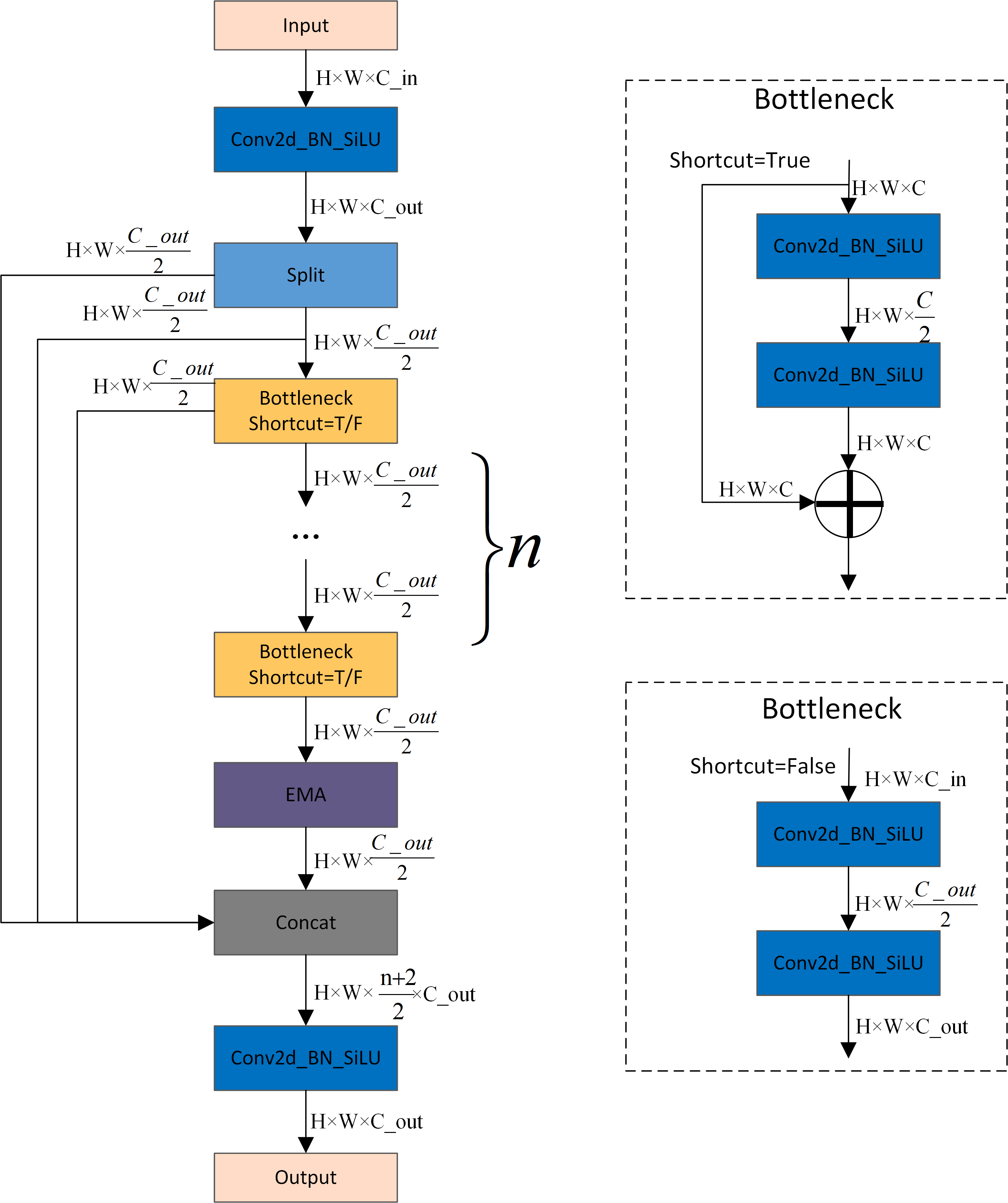}
    \caption{The structure of the C2f\_EMA module. In the Bottleneck module, T/F indicates whether a shortcut is used. T stands for true, and F stands for false.}
    \label{fig3:enter-label}
\end{figure}

In the YOLOv8 architecture, the C2f\_EMA model integrates multi-source local features using the EMA mechanism in the C2f backbone network, as shown in Figure.\ref{fig3:enter-label}. This parallel processing and self-attention strategy significantly improves performance, enhancing the model's accuracy, efficiency, and robustness, optimizing feature representation, and enabling it to excel in various visual tasks.

\subsection*{FasterPW}
To enhance the speed of neural networks, many studies have focused on reducing floating-point operations (FLOPs). However, the reduction in FLOPs does not always lead to a corresponding decrease in latency due to inefficient FLOPS. The FasterNeXt network reduces FLOPs while increasing FLOPS efficiency through PConv, reducing latency and enhancing computational speed without compromising accuracy. However, PConv has issues such as limited stride, insufficient receptive field, difficulty in determining the convolution ratio, and increased computational cost.

Our model replaces PConv with PWConv on the basis of the FasterNeXt network. PWConv is a special form of multi-channel convolution, where each convolution kernel is of size 1×1. It uses a 3D input feature map I of size (Hi × Wi × Ci) as input and a 4D filter F of size (1 × 1 × Ci × Co) to generate a 3D output feature map of size (Ho × Wo × Co), where Ho = Hi and Wo = Wi. 

The advantage of PWConv lies in its higher computational efficiency and reduced computational complexity. Specifically, PWConv serves as a local channel context aggrega-tor [44], utilizing pointwise channel interactions at each spatial location and reducing the number of channels to decrease computation.

\begin{figure}[htbp]
    \centering
    \includegraphics[width=0.5\linewidth]{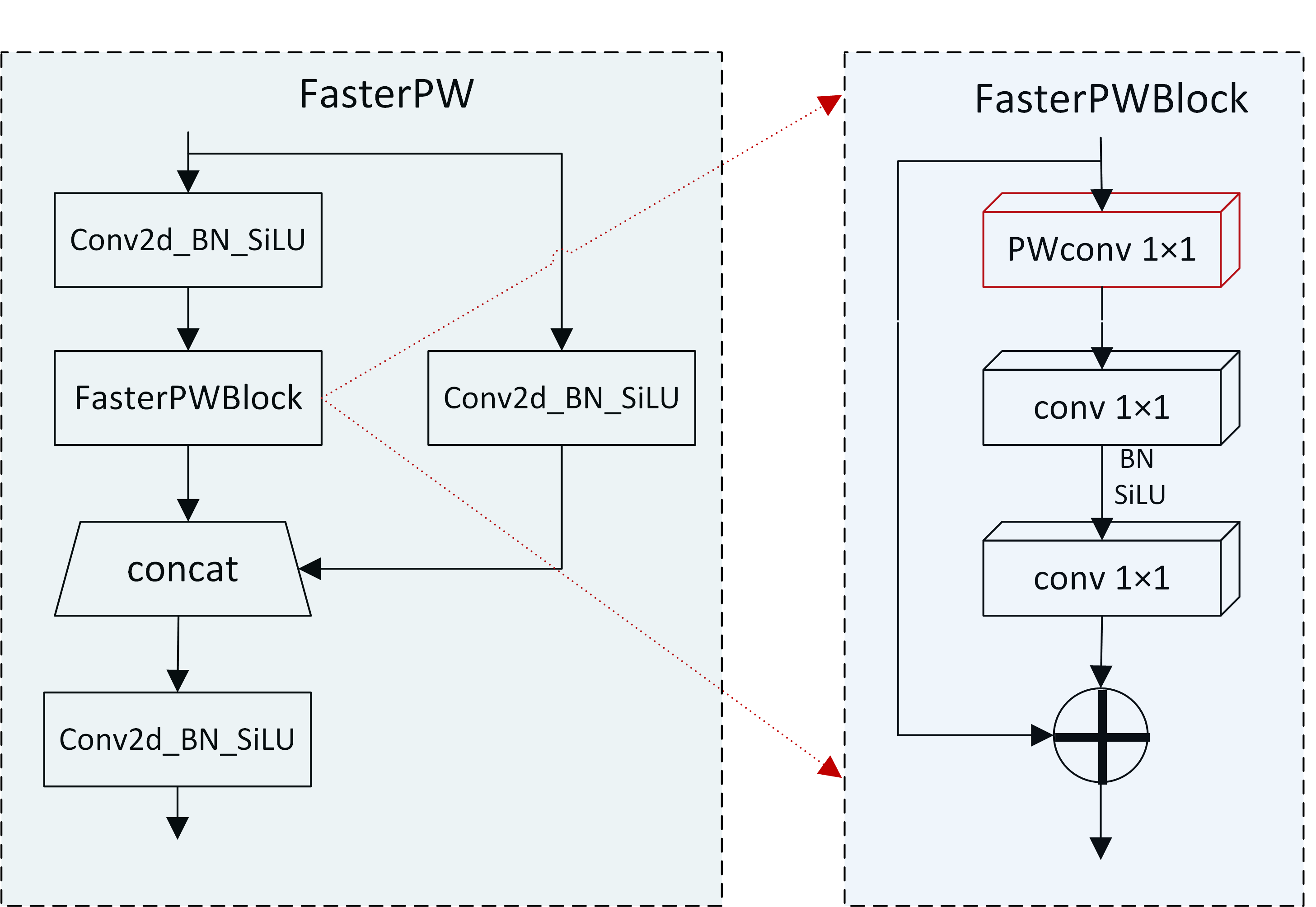}
    \caption{The construction process of the FasterPW structure. This structure adopts a lightweight design concept, initially applying PWConv to the FasterPWBlock network, then combining multiple FasterPWBlock and multiple ConvModules to construct an efficient and lightweight feature extraction network. }
    \label{fig4:enter-label}
\end{figure}

In this study, the core architecture of YOLOv8 underwent significant changes, with the C2f structure in its backbone network being replaced by the lightweight FasterPW series network, as shown in Figure.\ref{fig4:enter-label}. The FasterPW design includes three standard convolutional layers and FasterPWBlock module, forming a lightweight feature extraction network. This network strategically applies batch normalization (BN) \cite{MobileNetV2: Inverted Residuals and Linear Bottlenecks} and the SiLU activation function after pointwise convolutions to maintain feature diversity and reduce latency. SiLU has a stronger non-linear expression capability and efficient performance, making it the preferred activation function. By adopting the improved FasterNeXt network, we effectively reduced the number of network parameters, floating-point operations, and memory access times, achieving network lightweighting while maintaining efficient feature extraction capabilities \cite{MSA-YOLO: A Remote Sensing Object Detection Model Based on Multi-Scale Strip Attention}.
\subsection*{WFPN}
To enhance the feature extraction and fusion capabilities of fusion capabilities of the object detection model, we have adopted the concept of BiFPN, resulting in Weight\_Concat, which replaces the original Concat. The main principles of BiFPN are as follows:

•  Bidirectional Feature Fusion: BiFPN enables the fusion of features in both top-down and bottom-up directions, thus more effectively integrating features of different scales.
  
•   Weighted Fusion Mechanism: BiFPN optimizes the feature fusion process by assigning weights to each input feature, allowing the network to place greater emphasis on features with more substantial information content.

•   Structural Optimization: BiFPN optimizes cross-scale connections by removing nodes with only one input edge, adding additional edges between input and output nodes at the same level, and treating each bidirectional path as a feature network layer.

\begin{figure}[htbp]
    \centering
    \includegraphics[width=0.5\linewidth]{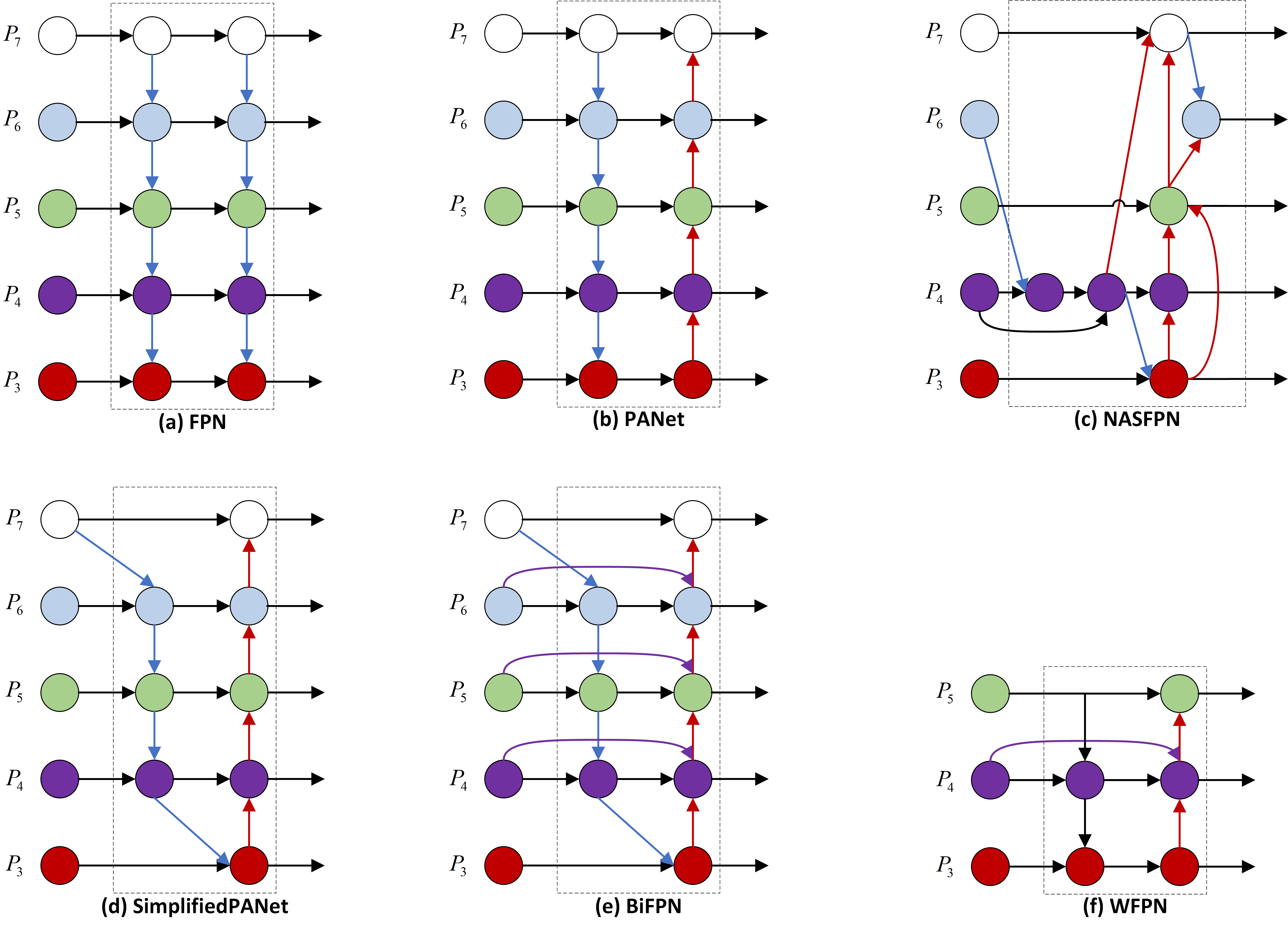}
    \caption{Comparison of different FPN structures, including the structures of \textbf{(a)} FPN, \textbf{(b)} PANet, \textbf{(c)} NASFPN, \textbf{(d)} SimplifiedPANet, \textbf{(e)} BiFPN and\textbf{ (f)}WFPN networks.}
    \label{fig5:enter-label}
\end{figure}

The six structures of FPN are shown in Figure.\ref{fig5:enter-label}, (a) is the original FPN feature fusion, which increases the high-level semantic information of large feature maps through the top-down branch, improving the detection accuracy of small objects, but at the same time, it makes the optimization of top-level features not as good as before, resulting in lower detection accuracy for large objects; (b) adds bottom-up feature fusion to (a), making the top-level features contain more detail information; (c) uses a feature fusion method searched by NAS. First, nodes with only one input edge are removed, as intuitively such nodes contribute less to the network's feature fusion, turning (b) into (d); secondly, if the output node and input node are at the same level, an edge is added between them, turning (d) into (e). (f) Combines the bidirectional feature fusion characteristics of (b) and the cross-scale linking features of (e) to form WFPN.

Weighted\_Concat enhances the semantic information of features by integrating effective bidirectional feature fusion, weighted fusion mechanisms, and structural optimization, as demonstrated in part (f) of Figure.\ref{fig5:enter-label}. Specifically, in terms of structural optimization, our method achieves cross-scale connections and differentiates input features of varying resolutions by introducing additional weights. This mechanism enables bidirectional cross-scale connections and rapid normalization fusion\cite{EfficientDet: Scalable and Efficient Object Detection}, allowing for more flexible adjustment of the influence of features at different resolutions, thereby better expressing the overall feature information.

\subsection*{CARAFE}
In multi-scale image labeling detection, feature upsampling is a key step. Traditional upsampling techniques often fail to fully utilize the semantic information in feature maps, limiting the effectiveness of feature fusion. Factorization methods leverage semantic information by learning upsampling kernels but increase the number of parameters and computational cost. Moreover, they use the same upsampling kernel at every position of the feature map, which cannot effectively utilize the semantic differences in the feature map. CARAFE has a larger receptive field, allowing for more effective aggregation of contextual information. Its upsampling kernel is closely related to the semantics of the feature map, effectively enhancing the multi-scale object detection performance after fusing multi-level features without significantly increasing parameters and computational cost. We replaced traditional upsampling with CARAFE in the neck part, further enhancing the performance of object detection, as shown in Figure.\ref{fig1:enter-label}.

\begin{figure}[htbp]
    \centering
    \includegraphics[width=0.5\linewidth]{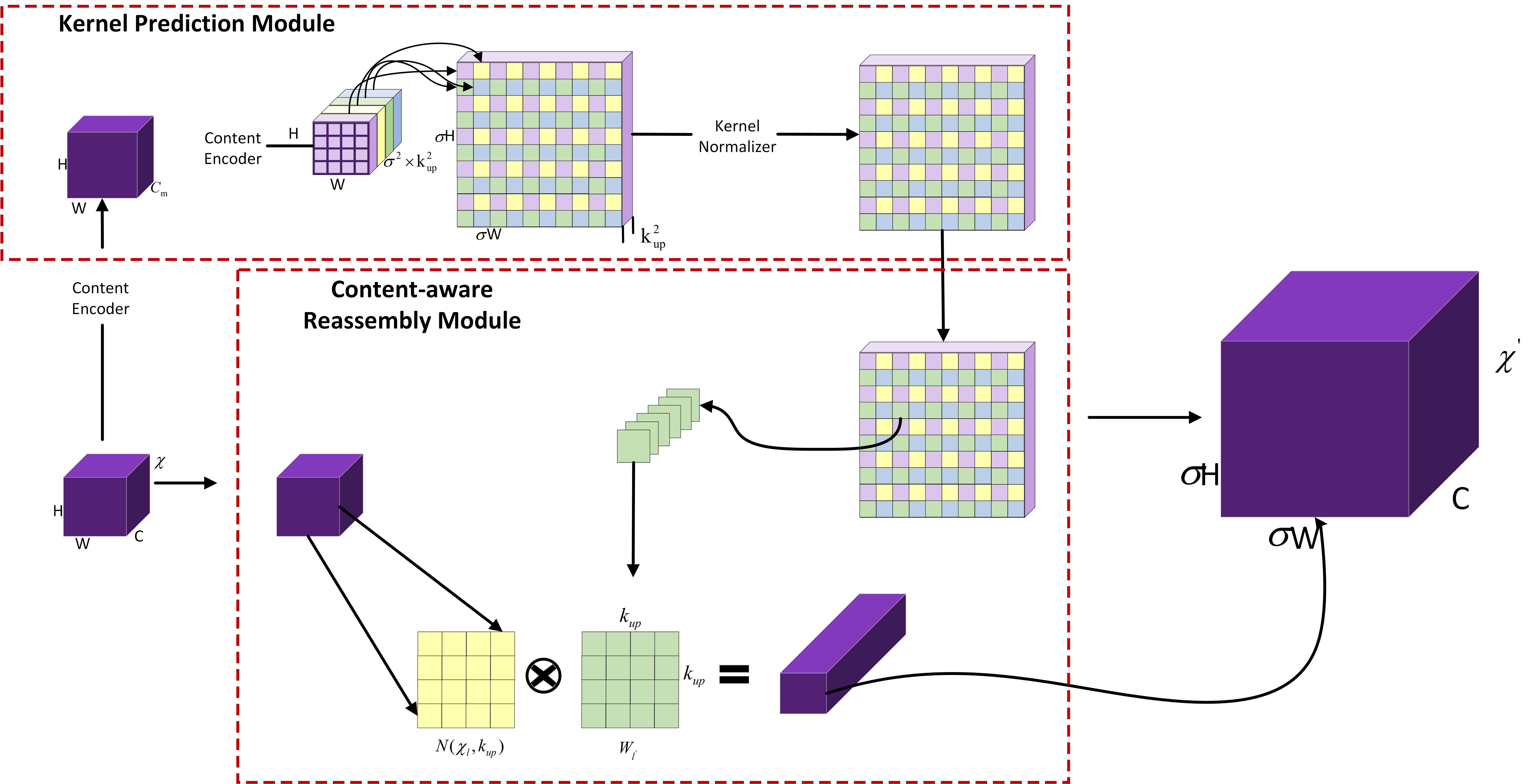}
    \caption{The overall framework of CARAFE, consisting of the kernel prediction module and content-aware assembly module.}
    \label{fig6:enter-label}
\end{figure}

CARAFE is a lightweight content-aware upsampling strategy that can accurately restore image details and reduce the loss of information in small objects. It consists of two parts: a kernel prediction module and a content-aware reassembly module \cite{Transformer-Based Object Detection with Deep Feature Fusion Using Carafe Operator in Remote Sensing Image}. The kernel prediction module predicts the upsampling kernels for each position, while the content-aware reassembly module uses these kernels to reassemble the feature maps pixel by pixel, achieving content-aware upsampling, as shown in Figure.\ref{fig6:enter-label}. This process makes the upsampling more perceptive and adaptive, improving the accuracy and effectiveness of upsampling.
CARAFE has the following innovative features:Firstly, the kernel prediction module $\psi$ predicts unique upsampling kernels $W_{l}'$ for each position $l'$ based on the neighboring locations of the input feature map $\chi_{l}$, achieving content adaptiveness. Secondly, the content-aware reassembly module uses these kernels for feature upsampling, effectively restoring detail information. This module contains a channel compressor, content encoder, and kernel normalizer, where the channel compressor compresses the input feature channels, the content encoder generates the reassembly kernels, and the kernel normalizer applies the softmax function. The content-aware reassembly module reassembles local area features through the weighted sum operator, allowing each pixel within the area to contribute differently to the upsampling pixels based on feature content rather than positional distance, enhancing the semantics of the feature map. Compared to the decomposition method, CARAFE has fewer parameters and lower computational complexity, and is more sensitive to details. Furthermore, the semantically enriched upsampling features generated by CARAFE significantly improve the model's detection and classification performance. In summary, CARAFE achieves efficient and semantically aware upsampling through content-adaptive kernel prediction and reassembly, making it an effective feature fusion module that can significantly enhance the model's classification and detection performance.

\section*{Experimental Details}\label{sec4}
The experimental environment parameters for this experiment are as follows. We used an Intel(R) Xeon(R) Gold 6248R @ 3.00 GHz processor and an NVIDIA GeForce RTX 3090 graphics card. The deep learning model framework used PyTorch 2.0.0 and Python 3.8, with CUDA version 11.7, and the operating system was Windows 11.

\subsection*{Benchmark Testing and Implementation Details}
\subsubsection*{Dataset}
\begin{figure}[htbp]
    \centering
    \includegraphics[width=0.5\linewidth]{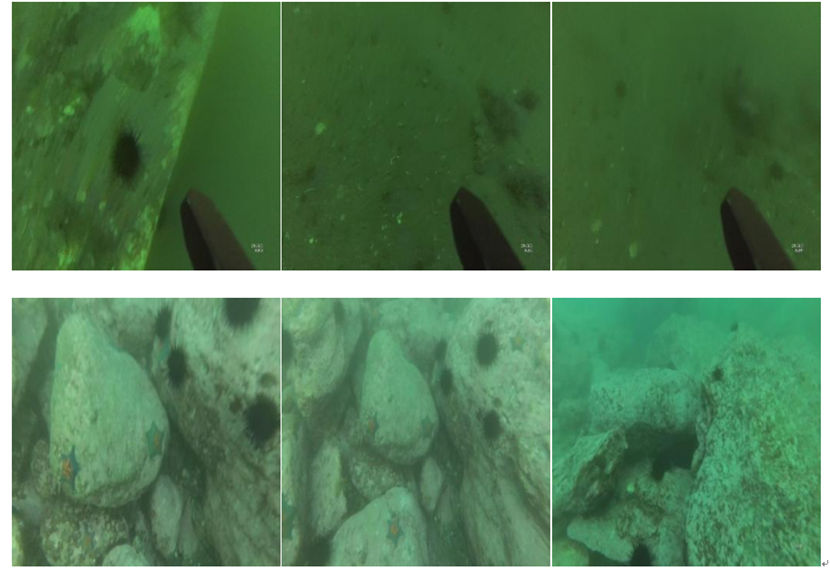}
    \caption{Example images from the URPC2019 dataset.}
    \label{fig7:enter-label}
\end{figure}

The dataset used in this paper is URPC2019, which is used to verify the effectiveness of our proposed model framework. URPC2019 is a publicly available dataset for underwater object detection, containing 5 different aquatic organism categories: echinoderms, starfish, holly, scallops, and seagrass, with a total of 3,765 training samples and 942 validation samples. Examples of dataset images are shown in Figure.\ref{fig7:enter-label}. Additionally, we conducted detection experiments on the URPC2020 dataset. Similar to URPC2019, URPC2020 is also an underwater dataset, but it differs in that it contains only four distinct categories: sea cucumbers, sea urchins, scallops, and starfish, with a total of 4,200 training samples and 800 validation samples. This further validates the feasibility of our model.
\subsubsection*{Parameter Settings}

To ensure the fairness and comparability of the model's effectiveness, we use official document as the pre-trained weight file for all experiments. At the same time, we employ the letterbox technique to adjust the input image size to 640×640, which allows the input image to retain the original aspect ratio while being adjusted to a fixed size, facilitating model training and inference. In the experiments, we set the number of iterations to 100, and some other important hyperparameters during the training phase of the model are shown in Table.\ref{tab1:hyperparameters}.

\begin{table}[htbp]
    \centering
    \caption{Some hyperparameters used in the training of EPBC-YOLOv8.}
    \label{tab1:hyperparameters}
    \begin{tabular}{cc} \toprule
        Parameters & Setup \\
        \midrule
        Epoch & 100 \\
        Batch size & 16 \\
        NMS IoU & 0.7 \\
        Image Size & 640×640 \\
        Initial Learning Rate & \(1 \times 10^{-2}\) \\
        Final Learning Rate & \(1 \times 10^{-2}\) \\
        \bottomrule
    \end{tabular}

\end{table}

\subsubsection*{Evaluation Criteria}

Intersection over Union (IOU) is a commonly used evaluation metric in object detection and image segmentation tasks. It measures the degree of overlap between the predicted bounding box (or segmentation result) and the true bounding box. IOU is defined by calculating the area of the intersection of the predicted bounding box and the true bounding box divided by the area of their union. See Equation.(\ref{equ7}) for details, where Intersection represents the area of the intersection of the predicted and true bounding boxes, and Union represents the area of their union. The value of IOU ranges from 0 to 1, with values closer to 1 indicating a higher degree of match between the predicted and true results, and values closer to 0 indicating a lower degree of match. 

\begin{eqnarray}\text{IoU} & = & \frac{\text{Area of Intersection}}{\text{Area of Union}}
\label{equ7}
\end{eqnarray}

In machine learning and statistics, False Positive (FP), True Positive (TP), False Negative (FN), and True Negative (TN) are common metrics used to evaluate the performance of classification models.

TP is the number of positive instances predicted as positive by the model, TN is the number of negative instances predicted as negative by the model, FP is the number of negative instances incorrectly predicted as positive by the model, and FN is the number of positive instances incorrectly predicted as negative by the model. The relationships between these four metrics are shown in Table.\ref{tab2:my_label}
Based on these metrics, other metrics such as Precision, Recall, Average Precision (AP), and Mean Average Precision (mAP) can be calculated.

\begin{table}[htbp]
    \centering
    \caption{The four relationships between predicted values and true values.}
    \label{tab2:my_label}
    \begin{tabular}{ccc}
        \toprule
        \diagbox{\textbf{Reference}}{\textbf{Prediction}} & \textbf{Positive} & \textbf{Negative}\\
        \midrule
        Positive & TP & FN \\
        Negative & FP & TN \\
        \bottomrule
    \end{tabular}

\end{table}

Precision is the proportion of true positives among the samples predicted as positive. High precision means that the model makes fewer misclassification judgments, but it does not guarantee that all positive instances are correctly identified, as shown in Equation.(\ref{equ8}).

\begin{eqnarray}\text{Precision} & = & \frac{TP}{TP + FP}
\label{equ8}
\end{eqnarray}

Recall is the proportion of true positives that are correctly predicted as positive among all actual positive instances. High recall means that the model can identify more positive instances, but it may also incorrectly predict some negative instances as positive, as shown in Equation.(\ref{equ9}).

\begin{eqnarray}\text{Recall} & = & \frac{TP}{TP + FN}
\label{equ9}
\end{eqnarray}

Average Precision is a metric used in tasks such as information retrieval and object detection. AP measures the ability of a model to rank results, and it is the average of the proportion of correct results returned. The higher the AP, the better the model is at ranking results, as shown in Equation.(\ref{equ10}).

\begin{eqnarray}AP & = & \int_{0}^{1} PR \, dr
\label{equ10}
\end{eqnarray}

Mean Average Precision is a metric used in tasks such as multi-class object detection. mAP is the average of the AP for each category, and it is used to evaluate the detection performance of the model across different categories, as shown in Equation.(\ref{equ11}).

\begin{eqnarray}\text{mAP} & = & \frac{1}{C} \sum_{i=1}^{C} AP_i
\label{equ11}
\end{eqnarray}

In our experiments, we use both precision and recall to evaluate the model's ability to classify positive and negative instances, and we use mAP to assess the model's performance in detecting and recognizing different categories of targets.

\subsection*{Comparative Experiments}

We conducted comparative experiments on the performance of our model EPBC-YOLOv8 and other models on the URPC2019 dataset, and the results are shown in Table.\ref{tab3:my_label}. It can be seen that our model has a significant reduction in both the number of parameters and FLOPs, which reduces the computational complexity of our model and is conducive to improving the accuracy of the model. The table shows that our model has a significant improvement in detection results compared to YOLOv5 and YOLOv8. Specifically, after modifications to the baseline YOLOv8n model, we observed an increase in mAP@0.5 precision by 2.3\%, which demonstrates the effectiveness of our model in optimizing YOLOv8.

\begin{table}[htbp]
    \caption{Comparison of the performance of the EPBC-YOLOv8 model with other algorithms on the URPC2019 dataset}
    \label{tab3:my_label}
        \begin{tabular}{ccccccccc}
            \toprule
            \multirow{2}{*}{Model} & \multicolumn{5}{c}{AP(\%)} & \multirow{2}{*}{mAP@0.5 (\%)} & \multirow{2}{*}{Para (M)} & \multirow{2}{*}{FLOPs(G)} \\
            \cmidrule{2-6}
            & Echinus & Starfish & Holothurian & Scallop & Waterweeds & & & \\
            \midrule
            Boosting RCNN\cite{Boosting R-CNN: Reweighting R-CNN samples by RPN’s error for underwater object detection}& 89.2 & 86.7 & 72.2 & 76.4 & 26.6 & 70.2 & 45.9 & 77.6 \\
            YOLOv5m& 92.0& 88.1& 75.2& 84.5& 24.2& 72.8& 20.9&47.9\\
            YOLOv5n& 91.9& 86.3& 58.4& 71.8& 17.6& 62.5& 1.8&4.2\\
            YOLOv5l& 92.4& 89.1& 73.6& 82.8& 36.6& 74.6& 46.2&108.3\\
            YOLOv5s & 92.4 & 89.3 & 74.7 & 83.8 & 28.4 & 73.7 & 7.0 & 16.0 \\
            YOLOv8s& 91.3& 89.0& 75.2& 84.9& 32.1& 74.5& 11.1&28.4\\
            YOLOv8m& 90.9& 89.5& 76.9& 85.7& 28.1& 74.2& 25.9&79.1\\
            YOLOv8l& 90.9& 90.4& 77.1& 84.8& 27.0& 74.0& 43.6&165.4\\
            YOLOv8n& 91.7& 89.2 & 76.1 & 82.8 & 32.3 & 74.4 & 3.0 & 8.2 \\
            EPBC-YOLOv8n & 92.1 & 89.3 & 74.5 & 82.0 & 45.5 & 76.7 & 2.6 & 7.6 \\
            \bottomrule
        \end{tabular}
\end{table}

\begin{figure}[htbp]
    \centering
    \includegraphics[width=0.5\linewidth]{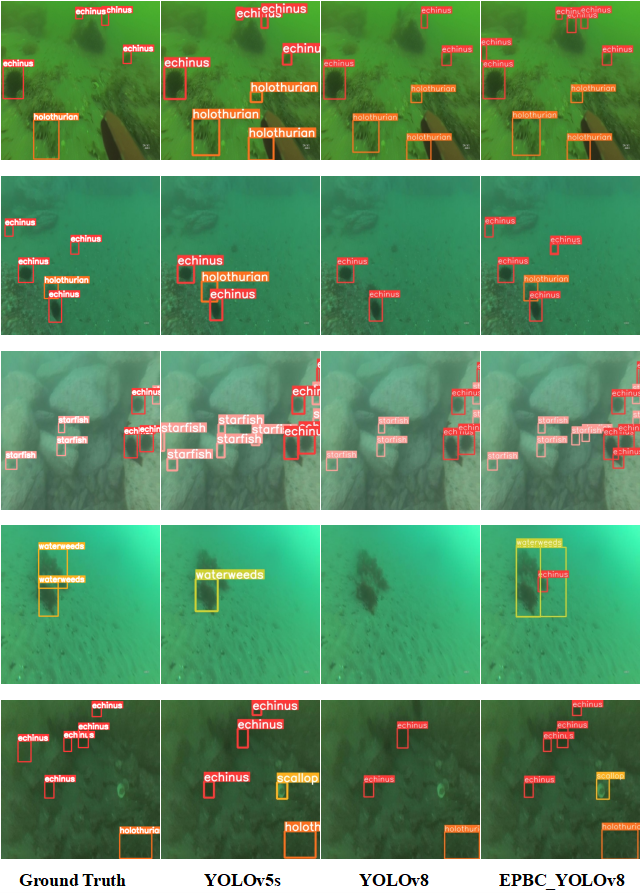}
    \caption{Comparison of object detection results between different models.}
    \label{fig8:enter-label}
\end{figure}

To highlight the effectiveness of our model, we conducted object detection on images from different scenes in the URPC2019 dataset. As shown in Figure.\ref{fig8:enter-label}, different models presented varying experimental results, with the four columns from left to right corresponding to Ground Truth, YOLOv5s, YOLOv8, and our optimized YOLOv8, respectively. By examining the charts, we can observe that, compared to YOLOv5s and YOLOv8, our model exhibits better detection performance. The first two rows demonstrate that the optimized YOLOv8 has a higher recall rate, while the last four rows reflect its higher accuracy. Therefore, comparing the detection results of our model with those of YOLOv5s and YOLOv8 further validates that our optimized model possesses superior detection performance. 
In addition, on the URPC2020 dataset, our model outperforms the original YOLOv8 and other models. As shown in Table.\ref{tab4:my_label}, compared to the YOLO series models, our model achieves an improvement of approximately 1\% in mAP@0.5. Although the improvement is not very significant, it represents a considerable advancement compared to other models. This indicates that our model possesses unique advantages on the URPC2020 dataset. 
\begin{table}[htbp]
    \centering
    \caption{Comparison of the performance of the EPBC-YOLOv8 model with other algorithms on the URPC2020 dataset}
    \label{tab4:my_label}
    \begin{tabular}{cccccl}
        \hline
        \multirow{2}{*}{Model} & \multicolumn{4}{c}{AP(\%)} & \multirow{2}{*}{mAP@0.5 (\%)} \\
        \cline{2-5}
        & Echinus & Starfish & Holothurian & Scallop & \\
        \hline
        YOLOv5x\cite{An Improved YOLOv5-Based Underwater Object-Detection Framework}& 67.5 & 87.9 & 75.1 & 81.4 &    78.0\\
        Faster-RCNN\cite{An Improved YOLOv5-Based Underwater Object-Detection Framework}& 57.5 & 77.7 & 51.4 & 74.5 & 65.3
\\
        Literature\cite{A Marine Organism Detection Framework Based on the Joint Optimization of Image Enhancement and Object Detection}& - & - & - & - & 69.4
\\
        YOLOv8& - & - & - & - & 78.3
\\
        EPBC-YOLOv8 & 69.7 & 88.5 & 74.7 & 83.0 & 79.0
\\
        \hline
    \end{tabular}
  
\end{table}

\subsection*{Ablation Study}
\subsubsection*{Effectiveness of C2f\_EMA in the Backbone}

Firstly, we modified the Bottleneck structure in the first C2f of the backbone part as shown in Figure .\ref{fig3:enter-label}.We attempted three optimization methods: C2f\_EMA is the optimized structure adopted in this paper; C2f\_FasterPW\_EMA is based on C2f\_EMA, with the Bottleneck in C2f replaced by FasterPW; C3\_FasterPW\_EMA applies the same operation to the C3 structure as C2f\_FasterPW\_EMA. As shown in Table.\ref{tab5:my_label}, the optimization of the C2f structure leads to an improvement in mAP@0.5, with the C2f\_EMA structure showing the most significant increase of 1.3\%.

\begin{table}[htbp]
    \centering
    \caption{Comparison of models after changing the first C2f in the backbone.}
    \label{tab5:my_label}
    \begin{tabular}{cc}
         \toprule
         Module & mAP@0.5 (\%) \\
         \midrule
         YOLOv8 & 74.4 \\
         YOLOv8+C2f\_EMA & 75.7 \\
         YOLOv8+C2f\_FasterPW\_EMA & 75.1 \\
         YOLOv8+C3\_FasterPW\_EMA & 75.1 \\
         \bottomrule
    \end{tabular}

\end{table}

\subsubsection*{Effectiveness of Each Module in EPBC-YOLOv8}

In this section, we take the original YOLOv8 as a basis and gradually add or remove components included in our model to explore the contribution of each component to the overall performance of the system model, thereby demonstrating their effectiveness in improving YOLOv8. A total of 15 ablation experiments were conducted, and the results are shown in Table.\ref{tab8:my_label}. By analyzing the table, we can see that different combinations of modules have different impacts on the performance  of the object detection system.

When using each of the four modules individually, there is an improvement compared to the original YOLOv8, with the increase in mAP in descending order being C2f\_EMA, CARAFE, WFPN, FasterPW . It can be seen that using the C2f\_EMA module alone results in the highest increase in mAP, which is 1.3\%.

When the modules are used in combination, there is an increase in mAP, and compared to using each module individually, the increase in mAP is generally larger. The best combination is when the C2f\_EMA, FasterPW, WFPN, and CARAFE modules are used simultaneously, achieving the highest mAP@0.5 of 76.7\%, which is a 2.3\% increase compared to the original YOLOv8.

In summary, based on the experimental results, the best performance improvement can be achieved by using the C2f\_EMA, FasterPW, WFPN, and CARAFE modules simultaneously. These results provide guidance for optimizing the design and configuration of object detection systems.

\begin{table}[htbp]
    \centering
    \caption{Proof of the effectiveness of each module in EPBC-YOLOv8.}
    \label{tab8:my_label}
    \begin{tabular}{ccccc}
         \toprule
         \multicolumn{4}{c}{Module} & \multirow{2}{*}{mAP@0.5 (\%)} \\
         \cmidrule{1-4}
         C2f\_EMA & FasterPW & WFPN & CARAFE & \\
         \midrule
         & & & & 74.4 \\
         \(\sqrt{}\) & & & & 75.7 \\
         & \(\sqrt{}\) & & & 74.6 \\
         & & \(\sqrt{}\) & & 74.8 \\
         & & & \(\sqrt{}\) & 75.3 \\
         \(\sqrt{}\) & \(\sqrt{}\) & & & 76.0 \\
         \(\sqrt{}\) & & \(\sqrt{}\) & & 75.8 \\
         \(\sqrt{}\) & & & \(\sqrt{}\) & 75.8 \\
         & \(\sqrt{}\) & \(\sqrt{}\) & & 75.1 \\
         & \(\sqrt{}\) & & \(\sqrt{}\) & 75.4 \\
         & & \(\sqrt{}\) & \(\sqrt{}\) & 76.0 \\
         \(\sqrt{}\) & \(\sqrt{}\) & \(\sqrt{}\) & & 76.2 \\
         & \(\sqrt{}\) & \(\sqrt{}\) & \(\sqrt{}\) & 76.5 \\
         \(\sqrt{}\) & & \(\sqrt{}\) & \(\sqrt{}\) & 76.6 \\
         \(\sqrt{}\) & \(\sqrt{}\) & & \(\sqrt{}\) & 76.4 \\
         \(\sqrt{}\) & \(\sqrt{}\) & \(\sqrt{}\) & \(\sqrt{}\) & 76.7 \\
         \bottomrule
    \end{tabular}

\end{table}

\subsection*{Result Analysis}

\begin{figure}[htbp]
    \centering
    \includegraphics[width=0.5\linewidth]{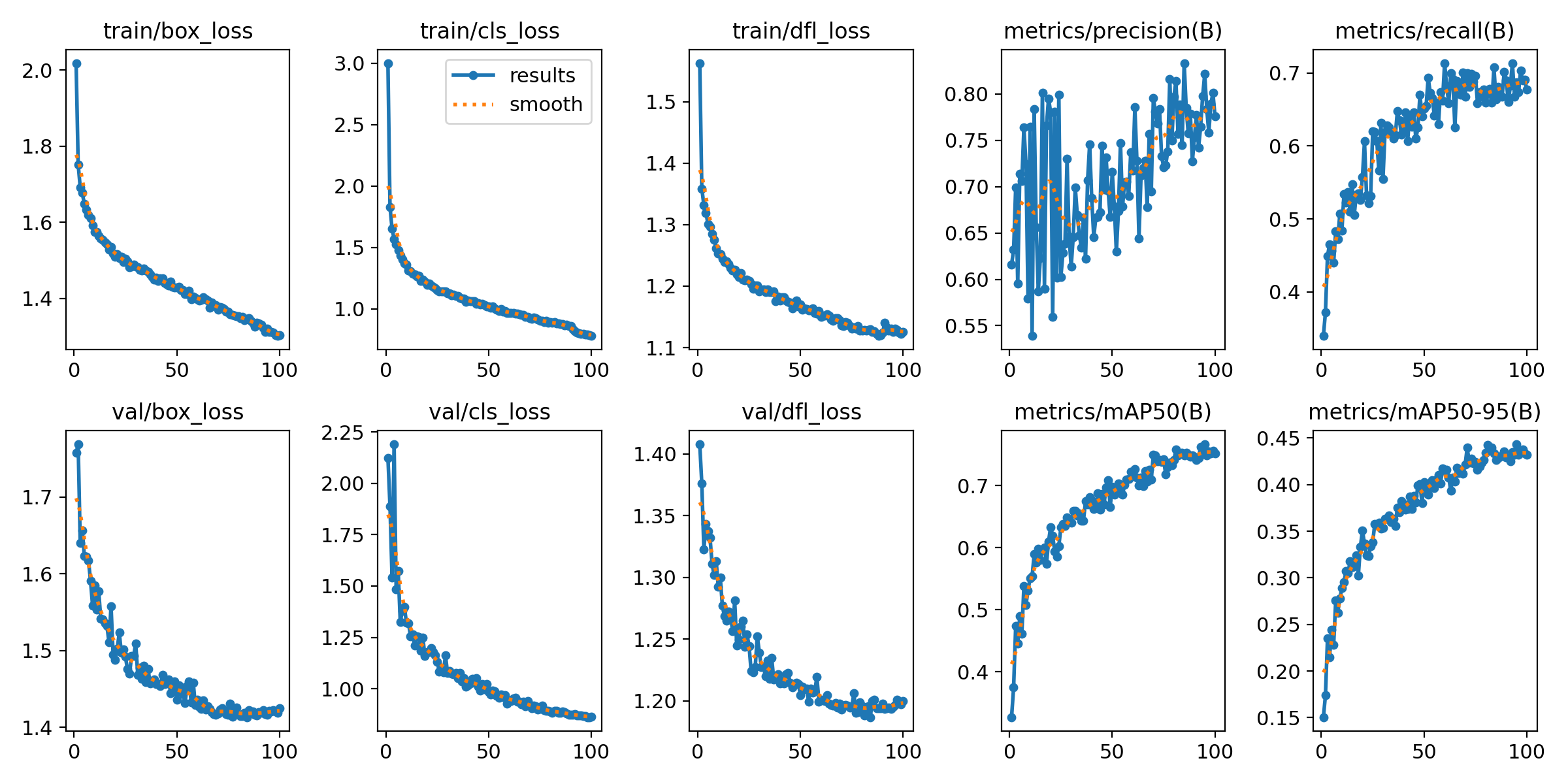}
    \caption{Trends of various parameters of the EPBC-YOLOv8 model over 100 epochs.}
    \label{fig9:enter-label}
\end{figure}

To verify the effectiveness of our improved EPBC-YOLOv8 model, we calculated the real-time changes in loss value, precision, recall, mAP@0.5, and mAP@0.95 for the EPBC-YOLOv8 model during 100 iterations on the training set and validation set. As can be seen from Figure.\ref{fig9:enter-label}, in both the training and validation sets, the real-time loss value of the EPBC-YOLOv8 model smoothly decreases with the increase of epochs and eventually converges. Especially in the validation set, the classification loss is more stable compared to the bounding box regression loss and the loss function for keypoint detection, indicating that the EPBC-YOLOv8 model has better classification performance for target categories. Simultaneously observing the changes in precision, recall, mAP@0.5, and mAP@0.5:0.95 values, all show an upward trend and good convergence, demonstrating that EPBC-YOLOv8 has good performance in object detection.

\begin{figure}[htbp]
    \centering
    \includegraphics[width=0.5\linewidth]{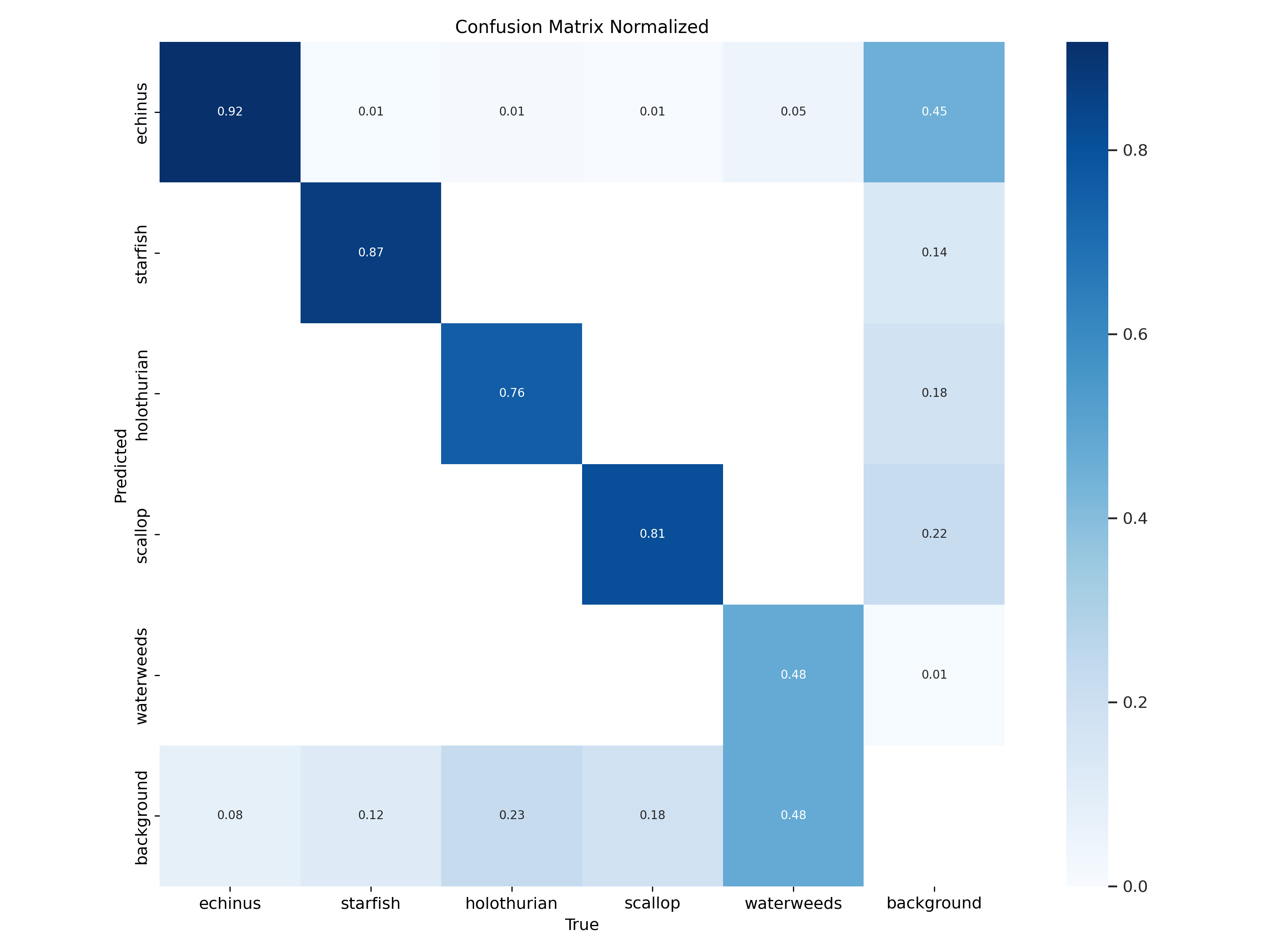}
    \caption{Confusion matrix of the EPBC-YOLOv8 model.}
    \label{fig10:enter-label}
\end{figure}

Figure.\ref{fig10:enter-label} shows the confusion matrix of our proposed EPBC-YOLOv8 model, which is normalized for a more intuitive display of the model's prediction performance. Each row of the matrix represents the actual category, while each column represents the predicted category. Observing the diagonal elements can intuitively reflect the prediction accuracy of the five categories in the URPC2019 dataset, and observing the off-diagonal elements can reveal the prediction situation between different categories. Therefore, we can conclude that the EPBC-YOLOv8 model has high prediction accuracy for most categories in the URPC2019 dataset, except for the more challenging category of waterweeds. The model's prediction error rate is relatively low, fully demonstrating the effectiveness of EPBC-YOLOv8 on this dataset.

\begin{figure}[htbp]
    \centering
    \includegraphics[width=0.5\linewidth]{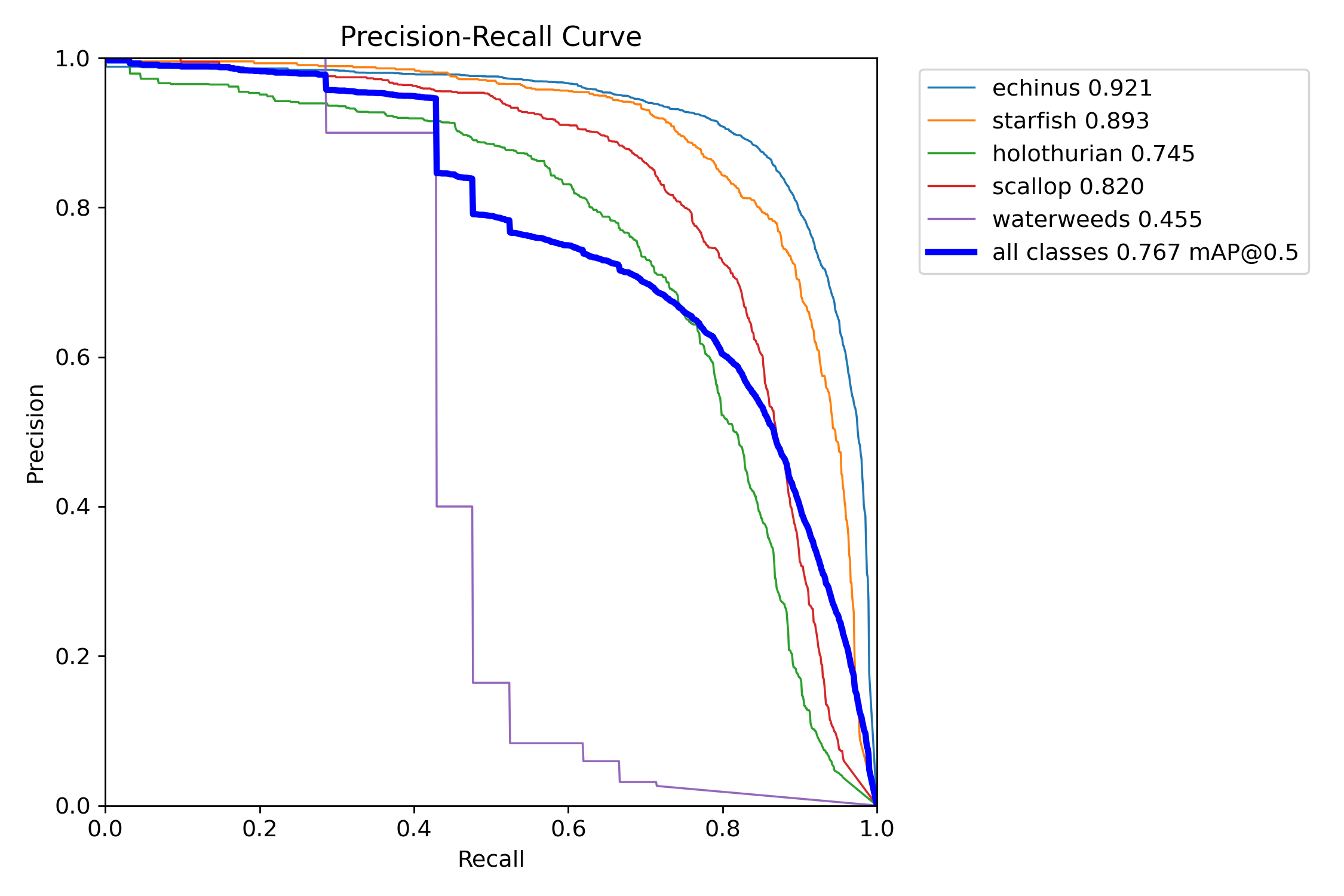}
    \caption{Precision-Recall (PR) curve of the EPBC-YOLOv8 model.}
    \label{fig11:enter-label}
\end{figure}

The Precision-Recall (PR) curve is a common method used to evaluate the performance of binary classifiers. In this curve, the horizontal axis represents recall, and the vertical axis represents precision. Precision and recall are two commonly used metrics for evaluating classifier performance. The PR curve shows the trade-off between precision and recall at different thresholds. Generally, we hope that the classifier has both high precision and high recall, so the closer the PR curve is to the top-right corner, the better the performance of the classifier. As can be seen in Figure.\ref{fig11:enter-label}., our model performs well, with the overall PR curve being closer to the top-right corner, indicating that our model has good performance.

\section*{Conclusions and Future Work}\label{sec5}

In this study, we propose EPBC-YOLOv8,an improved underwater organism detection framework, for complex underwater environments. In the backbone network of YOLOv8, we introduce an efficient multi-scale attention mechanism to focus on key regions of the image and extract strong features. The FasterPW part applies PWConv, using 1×1 dense convolution operations to fuse channel information and enhance computational accuracy. The neck module incorporates the content-aware characteristics of the CARAFE module, replacing traditional upsampling methods to enhance multi-scale object detection effects. The Concat connection of the model is augmented with the WFPN structure to improve performance and generalization ability. Experimental results demonstrate that EPBC-YOLOv8 achieves mAP@0.5 of 76.7\% and 79.0\% on the URPC2019 and URPC2020 datasets, respectively, which are 2.3\% and 0.7\% higher than the original YOLOv8.  Our improved method significantly outperforms the original YOLOv8 in terms of object detection accuracy, making it an efficient and practical underwater object detection framework. Future work will explore more advanced underwater image processing techniques to improve the accuracy of object detection in underwater environments.

\section*{Data availability}
Our code, model and dataset can be obtained from https://github.com/
zhuangxiting/EPBC-YOLOv8.git.



%\bibliography{your_external_BibTeX_file}

%=====================================
% References, variant B: internal bibliography
%=====================================

\section*{Author contributions statement}
Conceptualization, J.Z.; Methodology, X.Z. and X.J.; Software, J.C., X.J., and X.Z.; Validation, X.J., X.Z., and J.C.; Formal analysis, X.J.; Surveys, J.C.; Sourcing, J.Z.; Data curation, X.Z.; Writing-original draft preparation, X.J., X.Z., J.C., and Y.Z.; Writing-review and editing. J.Z.; Visualization, X.J.; Supervision, J.Z.; Project Management, J.Z.; Funding Acquisition, J.Z. All authors have read and agree to the published version of the manuscript.

\section*{Funding}
This research was funded by the college student innovation and entrepreneurship project of Hainnan University(Hdcxcyxm201704) and Hainan Provincial Natural Science Foundation of China (623RC449).

\section*{Competing interests}
Te authors declare no competing interests.

\section*{Additional information}
\textbf{Correspondence} and requests for materials should be addressed to J.Z.
\textbf{Reprints and permissions information} is available at www.nature.com/reprints.
\textbf{Publisher’s note }Springer Nature remains neutral with regard to jurisdictional claims in published maps and institutional afliations.

\textbf{Open Access} Tis article is licensed under a Creative Commons Attribution 4.0 International License, which permits use, sharing, adaptation, distribution and reproduction in any medium or format, as long as you give appropriate credit to the original author(s) and the source, provide a link to the Creative Commons licence, and indicate if changes were made. Te images or other third party material in this article are included in the article’s Creative Commons licence, unless indicated otherwise in a credit line to the material. If material is not included in the article’s Creative Commons licence and your intended use is not permitted by statutory regulation or exceeds the permitted use, you will need to obtain permission directly from the copyright holder. To view a copy of this licence, visit http://creativecommons.org/licenses/by/4.0/.

\end{document}